%% file: paper_arxiv.tex
\documentclass{article}

\usepackage[preprint]{neurips_2019}

\usepackage[switch]{lineno}

\usepackage[utf8]{inputenc} % allow utf-8 input
\usepackage[T1]{fontenc}    % use 8-bit T1 fonts
\usepackage{hyperref}       % hyperlinks
\usepackage{url}            % simple URL typesetting
\usepackage{booktabs}       % professional-quality tables
\usepackage{amsfonts}       % blackboard math symbols
\usepackage{nicefrac}       % compact symbols for 1/2, etc.
\usepackage{microtype}      % microtypography
\usepackage{natbib}
\usepackage[utf8]{inputenc}
\usepackage{algorithm}
\usepackage[noend]{algpseudocode}
\usepackage{algorithmicx}
\usepackage{amsmath,amssymb}
\usepackage{multirow}
\usepackage{multicol}

\usepackage{graphicx}

\usepackage{bm}
\usepackage{pifont} % symbols font

\newcommand{\eg}{\textit{e.g.}}
\newcommand{\ie}{\textit{i.e.}}

\DeclareMathAlphabet{\mathcal}{OMS}{cmsy}{m}{n}

\DeclareMathOperator*{\argmax}{arg\,max}

\def\algbackskip{\hskip-\ALG@thistlm}

\newcommand\MyBox[2]{
  \fbox{\lower0.75cm
    \vbox to 1.7cm{\vfil
      \hbox to 1.7cm{\hfil\parbox{1.4cm}{#1\\#2}\hfil}
      \vfil}%
  }%
}

\newcommand\MyBoxWide[2]{
  \fbox{\lower0.75cm
    \vbox to 1.7cm{\vfil
      \hbox to 3cm{\hfil\parbox{2.6cm}{#1\\#2}\hfil}
      \vfil}%
  }%
}

\newcommand{\KwInput}{\hspace*{\algorithmicindent} \textbf{Input:}}

\graphicspath{{figures/}}

\title{%Fairband: 
A Bandit-Based Algorithm for Fairness-Aware Hyperparameter Optimization}

\author{%
  André F. Cruz \\
  Feedzai \\
  \texttt{andre.cruz@feedzai.com} \\
    \And
    Pedro Saleiro \\
  Feedzai \\
  \texttt{pedro.saleiro@feedzai.com} \\
     \And
    Catarina Belém \\
  Feedzai \\
  \texttt{catarina.belem@feedzai.com} \\

        \And
    Carlos Soares \\
  Universidade do Porto \\
  \texttt{csoares@fe.up.pt} \\
  \And
    Pedro Bizarro \\
  Feedzai \\
  \texttt{pedro.bizarro@feedzai.com} \\

}

\begin{document}

\maketitle

%%%%%%%%%%%%%%
%% ABSTRACT %%
%%%%%%%%%%%%%%
% https://docs.google.com/document/d/1yY_4PUtfp4OrB6TXCSwm28eQNiRMjD3xejHzOHPLYsE/edit?usp=sharing

\begin{abstract}

Considerable research effort has been guided towards algorithmic fairness but there is still no major breakthrough. In practice, an exhaustive search over all possible techniques and hyperparameters is needed to find optimal fairness-accuracy trade-offs. Hence, coupled with the lack of tools for ML practitioners, real-world adoption of bias reduction methods is still scarce.
To address this, we present Fairband, a bandit-based fairness-aware hyperparameter optimization (HO) algorithm.
Fairband is conceptually simple, resource-efficient, easy to implement, and agnostic to both the objective metrics, model types and the hyperparameter space being explored.
Moreover, by introducing fairness notions into HO, we enable seamless and efficient integration of fairness objectives into real-world ML pipelines.
We compare Fairband with popular HO methods on four real-world decision-making datasets. We show that Fairband can efficiently navigate the fairness-accuracy trade-off through hyperparameter optimization. Furthermore, without extra training cost, it consistently finds configurations attaining substantially improved fairness at a comparatively small decrease in predictive accuracy.

\end{abstract}

\section{Introduction}

Artificial Intelligence (AI) has increasingly been used to aid decision-making in sensitive domains, including healthcare~\citep{Rajkomar2019}, criminal justice~\citep{berk2018fairness}, and financial services~\citep{board2017artificial}.
These algorithmic decision-making systems are accumulating societal responsibilities, often without human oversight. At the same time, Machine Learning (ML) models are usually optimized for a single global metric of predictive accuracy (\eg, binary cross-entropy loss on the training set), without taking into account possible side-effects and their real-world implications.

One potential side-effect is algorithmic bias, \ie, disparate predictive and error rates across sub-groups of the population, hurting individuals based on ethnicity, age, gender, or any other sensitive attribute~\citep{Angwin2016,Bartlett2019,Buolamwini2018}. This has several causes, from historical biases encoded in the data, to misrepresented populations in data samples, noisy labels, development decisions (\eg, missing values imputation), or simply the nature of learning under severe class-imbalance~\citep{Suresh2019}.

Algorithmic fairness~\citep{kleinberg2018algorithmic} is an emerging sub-field in AI that aims at reducing bias in decision-making systems.
Although a focus of extensive research in recent years, there are still no major breakthroughs in automatic bias reduction techniques~\citep{friedler2019comparative}.
Additionally, existing bias reduction techniques only target specific stages of the ML pipeline (\eg, data sampling, model training), and often only apply to a single fairness definition or family of ML models, limiting their adoption in practice.

There is a lack of practical methodologies and tools to seamlessly integrate fairness objectives and bias reduction techniques in existing real-world ML pipelines. As a consequence, treating fairness as a primary objective when developing AI systems is not standard practice yet \citep{saleiro2020dealing}.
Moreover, the absence of major breakthroughs in algorithmic fairness suggests that an exhaustive search over all possible ML models and bias reduction techniques may be necessary in order to find optimal fairness-accuracy trade-offs, hence discouraging AI practitioners.
% Moreover, the absence of major breakthroughs in algorithmic fairness suggests that testing several techniques and performing fairness-aware model selection may be a requirement for achieving optimal fairness-accuracy trade-offs. %may lead to better results than blindly applying a single technique.
%
However, in algorithmic fairness, model selection becomes a multi-criteria problem.
% If done naively, an exhaustive hyperparameter search over all possible ML models and bias reduction techniques may be computationally infeasible, hence discouraging AI practitioners.
We must provide fairness-aware AI development efficiently, with limited computational resources, minimizing predictive accuracy impact, and without the explicit need for expert knowledge. Therefore, this work contributes to the state-of-the-art in algorithmic fairness by bridging the gap between research and practice from an unexplored dimension: efficient fairness-aware hyperparameter optimization.
%Thereby, we overcome key shortcomings of methods from the literature: model dependence, metric dependence, and added complexity. 

\begin{figure}[t]
    \centering
    \includegraphics[width=0.6\columnwidth]{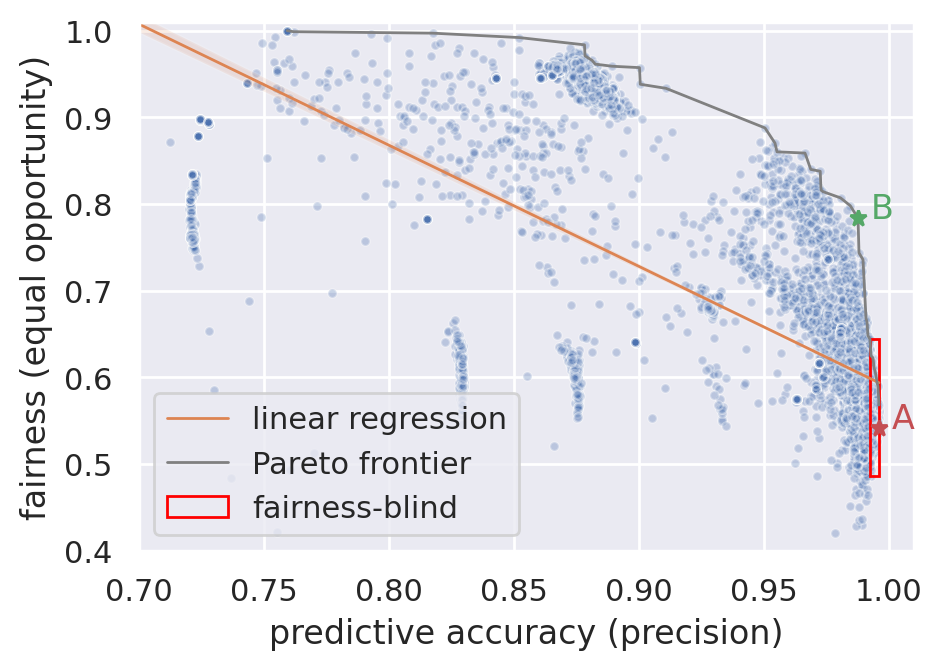}
    %\caption{
    %Fairness-Accuracy trade-off on the Adult dataset.
    %The top 10\% models with highest predictive accuracy are identified as the \textit{fairness-blind} region.}
    \caption{Fairness-accuracy trade-off of thousands of models on the Adult dataset. In orange, the linear regression relationship between accuracy and fairness; in the red rectangle, the top 10\% models with highest accuracy (the target region for a fairness-blind process); in light gray, the fairness-accuracy Pareto frontier; marked with an A, the model with highest accuracy; marked with a B, a model with 0.8\% lower accuracy and 44.8\% higher fairness than A, arguably a better trade-off, and one that would not be found by traditional fairness-blind techniques.
    }
    \label{fig:intro_tradeoff}
\end{figure}

Figure \ref{fig:intro_tradeoff} illustrates the fairness-accuracy trade-off over thousands of models trained on the Adult dataset~\citep{kohavi1996scaling}, each with a different hyperparameter configuration.
Currently employed model selection processes are fairness-blind, solely optimizing for predictive accuracy.
By doing so, these methods unknowingly target models with low fairness (region marked with a red rectangle).
However, as shown by the plotted Pareto frontier~\citep{Pareto1906}, we can achieve significant fairness improvements at small accuracy costs.
For instance, model B achieves 44.8\% higher fairness than model A (the model with highest predictive accuracy), at a cost of 0.8\% decrease in predictive accuracy, arguably a better trade-off.
While current fairness-blind techniques target model A, we target the region of optimal fairness-accuracy trade-offs to which model B belongs.
Indeed, we observe a large spread over the fairness metric at any level of predictive accuracy, even within this fairness-blind region.
Thus, it is absolutely possible to select fairer hyperparameter configurations without significant decrease in predictive accuracy.
With this in mind, we present Fairband, a bandit-based fairness-aware hyperparameter optimization algorithm.

Fairband is a resource-aware HO algorithm that targets optimal fairness-accuracy trade-offs, without increasing training budget or sacrificing parallelization.
% Fairband is able to find hyperparameter configurations attaining good fairness-accuracy trade-offs without having to increase training budget.
By making the hyperparameter search fairness-aware while maintaining resource-efficiency, we are enabling AI practitioners to adapt pre-existing business operations to accommodate fairness with controllable extra cost, and without significant friction.

The summary of our contributions are as follows:
\begin{itemize}

    \item Fairband, a flexible and efficient fairness-aware HO method for multi-objective optimization of the fairness-accuracy trade-off that is agnostic to both the explored hyperparameter space and the objective metrics (described in Section~\ref{chap:method}).

    \item A dynamic method to automatically search for good fairness-accuracy trade-offs without requiring manual weight parameterization (described in Section~\ref{sec:dynamic_alpha}).

    \item A competitive baseline for fairness-aware HO: random search with balanced fairness-accuracy ranking of hyperparameter configurations.

    \item Strong empirical evidence that hyperparameter optimization is an effective way to navigate the fairness-accuracy trade-off.

    \item Competitive results on 4 real world datasets: Fairband achieves 
    significantly improved fairness at a small predictive accuracy cost, and no extra budget when compared to literature HO baselines.
    
\end{itemize}

% %%%%%%%%%%%%%%%%%%%%%%%%%%%%%%%%%%%%%%%%%%%%%%%%%%%%%%%%%%%%%%%%%%%%%%%%%%%%%%%%%%%%%%%%%%%%%%%%%%%%%%%%%%%%%%%%
% % 
% %                                                  RELATED WORK 
% % 
% % %%%%%%%%%%%%%%%%%%%%%%%%%%%%%%%%%%%%%%%%%%%%%%%%%%%%%%%%%%%%%%%%%%%%%%%%%%%%%%%%%%%%%%%%%%%%%%%%%%%%%%%%%%%%%

\section{Related Work}

Algorithmic fairness research work can be broadly divided into three families: pre-processing, in-processing, and post-processing.

\textbf{Pre-processing} methods aim to improve fairness before the model is trained, by modifying the input data such that it no longer exhibits biases.
The objective is often formulated as learning new representations that are invariant to changes in specified factors (\eg, membership in a protected group)~\citep{calmon2017optimized,Creager2019,Edwards2016,Zemel2013}.
However, by acting on the data itself, and in the beginning of the ML pipeline, fairness may not be guaranteed on the end model that will be used in the real-world.

\textbf{In-processing} methods alter the model's learning process in order to penalize unfair decision-making.
The objective is often formulated as optimizing predictive accuracy under fairness constraints (or optimizing fairness under predictive accuracy constraints)~\citep{Cotter2018}.
Another approach optimizes for complex predictive accuracy metrics which include some fairness notion~\citep{zafar2017fairness}, akin to regularization.
However, these approaches are highly model-dependent and metric-dependent, and even non-existent for numerous state-of-the-art ML algorithms.

\textbf{Post-processing} methods aim to adjust an already trained classifier such that fairness constraints are fulfilled.
This is usually done by calibrating the decision threshold~\citep{Fish2016,Hardt2016}.
As such, these approaches are flexible and applicable to any score-based classifier.
However, one may argue that by acting on the model \textit{after} it was learned this process is inherently sub-optimal~\citep{Woodworth2017}. It is akin to knowingly learning a biased model and then correcting these biases, instead of learning an unbiased model from the start.

%% ------------ -----------------
%% Hyperparameter Optimization
%% ------------ -----------------
% Algorithmic fairness can be further achieved from a hyperparameter optimization (HO) perspective.
% Numerous methods have been proposed to address HO, including the traditionally employed algorithms, \textit{e.g.}, random-search, and recently introduced resource-aware algorithms, \textit{e.g.}, Successive Halving~\citep{Karnin2013} and Hyperband~\citep{Li2016b}.

Although a largely unexplored direction, algorithmic fairness can also be tackled from a hyperparameter optimization perspective.
%
% Numerous methods have been proposed to address HO, including the traditionally employed algorithms, \textit{e.g.}, random-search, and recently introduced resource-aware algorithms, \textit{e.g.}, Successive Halving~\citep{Karnin2013} and Hyperband~\citep{Li2016b}.
%
One of the simplest and most flexible HO methods is random search (RS).
This method iteratively selects combinations of random hyperparameter values and trains them on the full training set until the allocated budget %\footnote{The total amount of resources used (\eg, percentage of the training set used for training)}, $B$, 
is exhausted.
% Each hyperparameter configuration is trained on the full training set, and the best performing configuration is selected by evaluating on the validation set.
%
Although simple in nature, RS has several advantages that keep it relevant nowadays, including having no assumptions on the hyperparameter space, on the objective function, or even the allocated budget (\eg, it may run indefinitely).
Additionally, RS is known to generally perform better than grid search~\citep{Bergstra2012}, and to converge to the optimum as budget increases.
%
% Random search highly benefits from its arbitrariness, since a classifier's performance is not equally sensitive to all hyperparameters, and only a few are considerably impactful.
% In addition, random search has the advantage of compatibility with flexible resource allocation (as the the algorithm is not restricted to a pre-set number of evaluations, and instead may use, \eg, wall-clock time as the budget), as well as convergence to the optimum given enough resource.

\textbf{Bayesian Optimization} (BO) is a state-of-the-art HO method that consists in placing a prior (usually a Gaussian process) over the objective function to capture beliefs about its behavior~\citep{Shahriari2016}.
It iteratively updates the \textit{prior} distribution, $p(\lambda)$, using the evidence \textit{trial}, $X$, and forms a \textit{posterior} distribution of its behavior, $p(\lambda|X)$.
Afterwards, an acquisition function $a$ is constructed to determine the next query point $\lambda^{(i+1)}=\argmax_{\lambda}{a(\lambda)}$. This process is repeated in a sequential manner, continuously improving the approximation of the underlying objective function~\citep{Hutter2011a}.
%
%% Constrained BO %%

Previous work has extended BO to constrained optimization~\citep{cBO2014Kusner,cBO2014Snoek}, in which the goal is to optimize a given metric subject to any number of data-dependent constraints.
% In this setting, the acquisition function is weighed by the likelihood of fulfilling the constraints.
Recently, \cite{perrone2020fair} applied this approach to the fairness setting by weighing the acquisition function by the likelihood of fulfilling the fairness constraints.
However, constrained optimization approaches inherently target a single fairness-accuracy trade-off (which itself may not be feasible), leaving practitioners unaware of the possible fairness choices and their accuracy costs.
% A recent attempt at this \citep{perrone2020fair} corroborates that it is possible to navigate the fairness-accuracy trade-off through hyperparameter optimization.
%
Additionally, BO has been shown to scale poorly (cubically in the number of data points) and to lack behind bandit-based methods when under budget constraints%, achieving lackluster anytime performance
~\citep{Hutter2011a,Li2016b,Falkner2018b}.

\textbf{Successive Halving} (SH)~\citep{Karnin2013,Jamieson2016} casts the task of hyperparameter optimization as identifying the best arm in a multi-armed bandit setting.
Given a budget for each iteration, $B_i$, SH (1) uniformly allocates it to a set of arms (hyperparameter configurations), (2) evaluates their performance, (3) discards the worst half, and repeats from step 1 until a single arm remains.
Thereby, the budget for each surviving configuration is effectively doubled at each iteration.
SH’s key insight stems from extrapolating the rank of configurations’ performances from their rankings on diminished budgets (low-fidelity approximations).
However, SH itself carries two parameters 
% for whose values there is no clear method for setting: 
for which there is no clear choice of values: 
the total budget $B$ and the number of sampled configurations $n$.
%
% As the method leverages performance evaluations on a smaller budget (averaging $B/n$) to infer rank on a higher budget, 
We must consider the trade-off between evaluating a higher number of configurations (higher $n$) on an averaged lower budget per configuration ($B / n$), or evaluating a lower number of configurations (lower $n$) on an averaged higher budget.
The higher the average budget, the more accurate the extrapolated rankings will be, but a lower number of configurations will be explored (and vice-versa).
% Based on a stochastic learning process, SH may not be able to distinguish performance of different runs in the allocated budget for each configuration, which averages $B/n$. For a fixed $B$, the trade-off between choosing a low number of configurations (small $n$) with large average training time, or a higher number of configurations (larger $n$) with smaller training time must be considered. %
SH's performance has been shown to compare favorably to several competing bandit strategies from the literature~\citep{Audibert2010,Even-Bar2006,Jamieson2014,Kalyanakrishnan2012}.

\textbf{Hyperband} (HB)~\citep{Li2016b} addresses this ``$n$ \textit{versus} $B/n$'' trade-off by dividing the total budget into different instances of the trade-off, and then calling SH as a subroutine for each one.
This is essentially a grid search over feasible values of $n$.
HB takes two parameters $R$, the maximum amount of resources allocated to any single configuration; and $\eta$, the ratio of budget increase in each SH round ($\eta=2$ for the original SH).
%
% Lines 1--8 of Algorithm~\ref{algo:fairband} correspond to this mechanism of HB/SH.
%
Each SH run, dubbed a bracket, 
%(each corresponding to a different value of $s$), 
is parameterized by the number of sampled configurations $n$, and the minimum resource units allocated to any configuration $r$.
The algorithm features an outer loop that iterates over possible combinations of ($n$,$r$), and an inner loop that executes SH with the aforementioned parameters fixed.
The outer loop is executed $s_{max}+1$ times, $s_{max}=\lfloor log_\eta(R) \rfloor$, and the inner loop (SH) takes approximately $B$ resources.
Thus, the execution of Hyperband takes a budget of $(s_{max}+1)B$.
Table~\ref{tab:hyperband_brackets} displays the number of configurations and budget per configuration within each bracket when considering $\eta=3$ and $R=100$.

\begin{table}[t]    % Keep this table at the top of a page, it's important for future reference!
    \centering
    % \resizebox{\columnwidth}{!}{
    \input{tables/hyperband_brackets}
    % }
    \vspace{1em}
    \caption{Number of sampled configurations, $n_i$, and budget per configuration, $r_i$, for each Hyperband bracket (when the ratio of budget increase $\eta = 3$, and the maximum budget per configuration $R=100$). One budget unit equals 1\% of the train dataset.}
    \label{tab:hyperband_brackets}
\end{table}

Current bias reduction methods either 
(1) act on the input data and cannot guarantee fairness on the end model, 
(2) act on the model's training phase and can only be applied to specific model types and fairness metrics, %thus are inflexible with regards to the chosen model or metric, 
or (3) act on a learned model's predictions and thus are limited to act on a sub-optimal space.
%
% In practise, the choice of which bias reduction method to employ is usually done by so-called ``graduate student descent''.
% In practise, a large selection of bias reduction methods has to be evaluated in order to find optimal fairness-accuracy trade-offs.
%
On the other hand, hyperparameter optimization is simultaneously model independent, metric independent, and already a componeent of existing real-world ML pipelines.
%
% HO can supplement current bias reduction methods by including these in the search space and automatically selecting which method to employ.
%
By introducing fairness objectives on the HO phase in a effient way, we aim to help real-world practitioners to seamlessly finding optimal fairness-accuracy tradeoffs regardless of the underlying model type or bias reduction method.

%%%%%%%%%%%%%%%%%%%%%%%%%%%%%%%%%%%%%%%%%%%%%%%%%%%%%%%%%%%%%%%%%%%%%%%%%%%%%%%%%%%%%%%%%%%%%%%%%
% 
%                                                  METHOD / PROPOSED SOLUTION 
% 
%%%%%%%%%%%%%%%%%%%%%%%%%%%%%%%%%%%%%%%%%%%%%%%%%%%%%%%%%%%%%%%%%%%%%%%%%%%%%%%%%%%%%%%%%%%%%%%%%
\section{Fairband}
\label{chap:method}

%% Proposal:
% Black-box algorithmic fairness
Our stated goal is to enable flexible and efficient fairness-aware hyperparameter optimization.
To this end, we present Fairband (FB), a novel bandit-based algorithm for multi-criteria hyperparameter optimization.

By acting on the algorithms' hyperparameter space, we benefit from the advantages of in-processing bias reduction methods, while avoiding its shortcomings (\eg, dependency on the metrics and model).
That is, we aim to improve model fairness by acting on the models' training phase, instead of simply correcting the data (pre-processing), or correcting the predictions (post-processing).
Nonetheless, as any multi-objective optimization problem with competing metrics, we can only aim to identify good fairness-accuracy trade-offs~\citep{zafar2017pareto}.
The decision on which trade-off to employ should be left to the model's stakeholders.

Aiming to benefit from the efficiency of state-of-the-art resource-aware HO methods, we build our method on top of Successive Halving (SH)~\citep{Karnin2013} and Hyperband (HB)~\citep{Li2016b}.
% We propose introducing fairness objectives into the state-of-the-art resource-aware hyperparameter optimization methods: Successive Halving~\citep{Karnin2013} (SH) and Hyperband~\citep{Li2016b}.
Thus, Fairband benefits from these methods' advantages: being both model- and metric-agnostic, having efficient resource usage, and trivial parallelization.
% Furthermore, these methods have a high capacity for exploring the hyperparameter space.
Furthermore, these methods are highly exploratory and therefore prone to inspect broader regions of the hyperparameter space.
For instance, in our experiments, Hyperband evaluates approximately six times more configurations than Random Search with the same budget\footnote{With the parameters used on the Hyperband seminal paper~\citep{Li2016b}, it evaluates 128 configurations compared with 21 configurations evaluated by Random Search on an equal budget.}.
Most importantly, these algorithms are easily extendable: by changing the sampling strategy, by changing how we evaluate a given hyperparameter configuration, or by changing how we select the top configurations to be kept between iterations.
As an example, BOHB~\citep{Falkner2018b} is an extension of Hyperband that introduces Bayesian optimization into the sampling strategy.

%%
%% PENALIZING UNFAIRNESS
%%
In order to introduce fairness objectives into the hyperparameter optimization process, we assume our goal is the maximization of a predictive accuracy metric $a$ and a fairness metric $f$, for $a, f \in [0, 1]$ 
(or, equivalently, the minimization of $1 - a$ and $1 - f$).

Our method weighs the optimization metric by both predictive accuracy and fairness, parameterized by the relative importance of predictive accuracy $\alpha \in [0, 1]$ (see Equation \ref{eq:weighted_combination}).
% Note that any number of different averaging functions could be employed at this step (\eg, arithmetic, harmonic, or geometric mean).
This is a popular method for multi-objective optimization known as \textit{weighted-sum scalarization}~\citep{Deb2014}.
%
% Note that any number of different combinations or averaging functions could be employed at this step (\eg, arithmetic, harmonic, or geometric mean) but all rely on the same hypothesis: if model $m_1$ represents a better fairness-accuracy trade-off than model $m_2$ with a short training budget, then this distinction is likely to be maintained with higher training budget.
By employing this technique in a bandit-based setting, we rely on the hypothesis that if model $m_a$ represents a better fairness-accuracy trade-off than model $m_b$ with a short training budget, then this distinction is likely to be maintained with a higher training budget.
% Some models are fairer than others, we hypothesise this relative distinction is maintained as budget increases, akin to low-fidelity predictive accuracy estimates
Thus, by selecting models based on both fairness and predictive accuracy, we are guiding the search towards fairer and better performing models.
These low-fidelity %(but always better than random, $p>0.5$)
estimates of future metrics on lower budget sizes is what drives Hyperband and SH's efficiency in hyperparameter search.
Our proposed optimization metric, $o$, is given by the following equation:

\begin{equation}
\label{eq:weighted_combination}
    o = \alpha \cdot \textit{a} + (1 - \alpha) \cdot \textit{f}
\end{equation}

Accordingly, all models are evaluated in both fairness and predictive accuracy metrics on a holdout validation set.
Computing fairness does not imply significant extra computational cost, as it is based on the same predictions used to estimate predictive accuracy.
Additionally, fairness assessment libraries are readily available~\citep{Saleiro2018}.

\subsection{Dynamic $\alpha$}
\label{sec:dynamic_alpha}

% Our proposed value of $\alpha$ is calculated
As a multi-objective optimization problem, we aim to identify configurations that represent a balanced trade-off between the two target metrics, $a, f \in [0, 1]$.
Simply employing $\alpha = 0.5$ would assume one unit of improvement on one metric is effectively equal to one unit of improvement on the other, %which is not the case on most real-world problems.
and would guide the search towards a single region of the fairness-accuracy trade-off.
% which is highly dependent on the chosen metrics and the real-world setting in which the model will be deployed.

%
Thus, aiming for a complete out-of-the-box experience without the need for specific domain knowledge, we propose a heuristic for automatically setting $\alpha$ values targeting a broader exploration of the Pareto frontier~\citep{Pareto1906} and a balance between searching for fairer or more accurate configurations.
We dub this variant \textit{FB-auto}.
Assuming that $\alpha$ values can indeed guide the search towards different regions of the fairness-accuracy trade-off (which we will empirically see to be true), our aim is to efficiently explore the Pareto frontier in order to find a comprehensive selection of balanced trade-offs.
As such, if our currently explored trade-offs correspond to high accuracy but low fairness, we want to guide the search towards higher fairness (by choosing a lower $\alpha$).
Conversely, if our currently explored trade-offs correspond to high fairness but low accuracy, we want to guide the search towards higher accuracy (by choosing a higher $\alpha$).
%
% Hence, we propose computing a different value of $\alpha$ at each Fairband iteration, given the average model fairness $\overline{f}$ and average model predictive accuracy $\overline{a}$ at that iteration.

%%%%%%%%%%%%%%%%%%%%%%%%%%%%%%%
%% Dynamic-alpha Explanation %%
%%%%%%%%%%%%%%%%%%%%%%%%%%%%%%%
To achieve the aforementioned balance we need a proxy-metric of our target direction of change.
% To preserve the semantic meaning of $\alpha$, that is, the relative importance of the accuracy metric at a given iteration, we compute a proxy measure of how many better-performing models are necessary, on average, to balance the two target metrics.
This direction is given by the difference, $\delta$, between the average model fairness, $\overline{f}$, and average predictive accuracy, $\overline{a}$, as shown in Equation~\ref{eq:delta}:

\begin{equation}
\label{eq:delta}
    \delta = \overline{f} - \overline{a}, \ \delta \in \left[-1, 1\right]
\end{equation}

Hence, when this difference is negative, $\overline{f} < \overline{a}$, the models we sampled thus far tend towards better-performing but unfairer regions of the hyperparameter space.
Consequently, we want to decrease $\alpha$ to direct our search towards fairer configurations.
Conversely, when this difference is positive, $\overline{f} > \overline{a}$, we want to direct our search towards better-performing configurations, increasing $\alpha$.
% As  $f, a \in \left[0, 1\right]$ for every model, $\delta$ is always bound by the closed-interval $\left[-1, 1\right]$.
% Hence, the value of $\alpha$ should change proportionally to $\delta$, by some constant $c \in \mathbb{R}^{+}$.
We want this change in $\alpha$ to be proportional to $\delta$ by some constant $k > 0$, such that

\begin{equation}
\label{eq:alpha_sigma}
 \frac{d \alpha}{d \delta} = k, \ k \in \mathbb{R}^{+}
\end{equation}

and by integrating this equation we get
\begin{equation}
\label{eq:integrated_alpha}
\alpha = k \cdot \delta + c, \ c \in \mathbb{R}
\end{equation}

with $c$ being the constant of integration.
Given that $\delta \in \left[-1, 1\right]$, and together with the constraint that $\alpha \in \left[0, 1\right]$, the only feasible values for $k$ and $c$ are $k=0.5$ and $c=0.5$.
Hence, the computation of dynamic-$\alpha$ is given as follows by Equation~\ref{eq:dynamic_alpha}:
% as $\alpha = 0.5$ corresponds to a balance between accuracy and fairness metrics,
% we want to increase the value of $\alpha$ up to $\alpha = 1$ when targeting better-performing configurations, and decrease it up to $\alpha = 0$ when targeting fairer configurations.
% Additionally, we want this change in $\alpha$ to be proportional to $\delta$ by some constant $c \in \mathbb{R}^{+}$ (see Equation~\ref{eq:alpha_derivative}).
%
% Hence, after scaling down $\delta$ to the interval $\left[-0.5, 0.5\right]$, the computation of dynamic-$\alpha$ is given as follows by Equation~\ref{eq:dynamic_alpha}.
% Lastly, we apply min-max scaling to $\delta$ to scale it down to the interval $\left[0, 1\right]$ (see Equation ~\ref{eq:min_max_scale}). Overall, for a maximization setting, the computation of the dynamic-$\alpha$ can be described by Equation~\ref{eq:dynamic_alpha}. 

% \begin{equation}
% \label{eq:min_max_scale}
%     \alpha = \frac{(\overline{f} - \overline{a}) - (-1)}{1 - (-1)}
% \end{equation}

% \begin{equation}
% \label{eq:alpha_derivative}
%     \frac{d \alpha}{d \delta} = c \Rightarrow
%     % \alpha = \int{c \, d\delta} \Rightarrow
%     \alpha = k + c \cdot \delta, \ k \in \mathbb{R}%, c \in \mathbb{R}^{+}
% \end{equation}

\begin{equation}
\label{eq:dynamic_alpha}
    \alpha = 0.5 \cdot (\overline{f} - \overline{a}) + 0.5
\end{equation}

% Figure~\ref{fig:dynamic_alpha} shows a plot of Equation~\ref{eq:dynamic_alpha}.
% Note that to adjust this to a minimization setting, a simple change would be required: $\alpha = 0.5 + (\overline{a} - \overline{f}) / 2$.

% This dynamic $\alpha$ calculation is guided by the fact that the predictive accuracy and fairness metrics may have different ranges of values, and these ranges are likely to vary between Fairband iterations (as the search progresses).
%
Moreover, earlier iterations are expected to have lower predictive accuracy (as these are trained on a lower budget), while later iterations are expected to have higher predictive accuracy.
By computing new values of $\alpha$ at each Fairband iteration, we promote a dynamic balance between these metrics as the search progresses, predictably giving more importance to accuracy on earlier iterations but continuously moving importance to fairness as accuracy increases (a natural side-effect of increasing training budget).
% By computing new values of $\alpha$ at each Fairband iteration, we promote the exploration of different regions of the fairness-accuracy space, instead of focusing our search on a single trade-off dictated by a single value of alpha.
%
% Thus, if earlier iterations have lower predictive accuracy (which is expected), but higher fairness, then the search would be guided towards higher performing models.
% Conversely, if later iterations have higher predictive accuracy, but lower fairness, then search is guided towards fairer models, promoting a balance between both metrics.
%
% Additionally, as no information is shared between SH brackets, Fairband maintains effortless parallelization.
%
Thus, we enable parameter-free fairness-aware optimization via \textit{dynamic} $\alpha$.%, as well as, a fixed fine-tuning to specific fairness-accuracy trade-offs via \textit{static} choices of $\alpha$.

\subsection{Algorithm}
\label{sec:algorithm}

\begin{algorithm}[th]
\caption{Fairband}
\label{algo:fairband}
\KwInput~maximum budget per configuration $R$, \\
\hphantom{\KwInput} $\eta$ (default $\eta = 3$), \\
\hphantom{\KwInput} $\alpha$ (default $\alpha = \textit{auto}$)
\begin{algorithmic}[1]
\State $s_{max} \gets \lfloor\log_{\eta}{(R)}\rfloor$ \Comment{define number of brackets}
\State $B \gets (s_{max} + 1)\cdot{R}$      \Comment{compute budget per bracket}
\For{$s \in \{s_{max}, s_{max} - 1, ..., 0\}$}      \Comment{iterate through SH brackets, as per \cite{Li2016b}}
    \State $n \gets \lceil \frac{B}{R} \cdot \frac{\eta^{s}}{s+1} \rceil$, $r \gets R\cdot\eta^{-s}$      \Comment{choice of $n$ \textit{versus} $B/n$ trade-off}
    \State $T \gets \textit{get\_hyperparameter\_configurations}(n)$

    \For{$i \in \{0, ..., s\}$}             \Comment{run Successive Halving}
        \State $n_i \gets \lfloor{n}\cdot{\eta^{-i}}\rfloor$    \Comment{train $n_i$ configurations}
        \State $r_i \gets r\cdot\eta^{i}$       \Comment{$r_i$ training budget per config.}
        \State $M \gets \{\textit{train\_with\_budget}(\bm\lambda, r_i) : \bm\lambda \in T\}$
        \State $A \gets \{\textit{evaluate\_accuracy}(m_{\bm\lambda}) : m_{\bm\lambda} \in M\}$
        \State $F \gets \{\textit{evaluate\_fairness}(m_{\bm\lambda}) : m_{\bm\lambda} \in M\}$

        \If{$\alpha = \textit{auto}$}                                \Comment{compute dynamic $\alpha$ if applicable}
            \State $\overline{f} \gets$ \textit{sum}$(F) / |F|$        \Comment{average fairness}
            \State $\overline{a} \gets$ \textit{sum}$(A) / |A|$        \Comment{average accuracy}
            \State $\alpha \gets 0.5 \cdot (\overline{f} - \overline{a}) + 0.5 $           %\Comment{assumes maximization}
        \EndIf

        \State $O \gets \{\alpha \cdot A[m_{\bm\lambda}] + (1 - \alpha) \cdot F[m_{\bm\lambda}] : m_{\bm\lambda} \in M\}$          \Comment{compute objective metric, $o$}
        \State $I \gets \textit{argsort}(O)$                                        \Comment{sorted in descending order}
        \State $k \gets \lfloor{n_i / \eta}\rfloor$         \Comment{number of configurations to keep}
        \State $T \gets T[I[0: k]]$                          \Comment{select top $k$ configurations}
    \EndFor
\EndFor
\State \Return $\bm\lambda^{*}$, configuration with maximal intermediate goal seen so far
\end{algorithmic}
\end{algorithm}

Firstly, we consider 
%the definition of 
a broad hyperparameter space as a requirement for the effective execution of Fairband.
%Crucially, we 
We consider as hyperparameters any decision in the ML pipeline, as bias can be introduced at any stage of this pipeline~\citep{barocas2016big}.
Thus, an effective search space includes which model type to use, the model hyperparameters which dictate how it is trained, and the sampling hyperparameters which dictate the distribution and prevalence rates of training data.

Fairband is detailed in Algorithm~\ref{algo:fairband}.
% At a high level, the method is based on Hyperband, and calls SH as a subroutine.
It has three parameters: 
%It is provided three parameters: 
$R$, the maximum amount of resources allocated to any single configuration; 
$\eta$, which dictates both the budget increase and the proportion of configurations discarded in each SH round; 
and $\alpha$, the relative importance of predictive accuracy versus fairness. 
The first two parameters are ``inherited'' from Hyperband. On the other hand, 
%The latter parameter, 
$\alpha$ may be omitted, relying on our proposed dynamic computation of $\alpha$.

Our method incorporates Hyperband's exploration of SH's brackets, iterating 
%Following the same schema of SH brackets as Hyperband, our method iterates 
through different parameters of SH corresponding to different instances of the ``n versus B/n'' trade-off (whether to evaluate more configurations on a lower budget, or less configurations on a higher budget).
For each SH run (or bracket), our method (1) randomly samples $n$ hyperparameter configurations, (2) trains each sampled configuration on the allocated budget $r_i$, (3) evaluates their predictive accuracy and fairness, and (4) selects the top $k$ configurations on the objective metric $o$.
Optionally, between steps 3 and 4, a dynamic value of $\alpha$ will be computed as described in Section~\ref{sec:dynamic_alpha}. %from the average fairness and predictive accuracy at that iteration.
Note that when using a static value of $\alpha=1$ our method is functionally equivalent to the traditional Hyperband algorithm (only taking predictive accuracy into account).

%% ON SELECTION ALPHA
The result of our method's execution is a collection of hyperparameter configurations that effectively represent the fairness-accuracy trade-off.
One could plot all available choices on the fairness-accuracy space and manually pick a trade-off, according to whichever business constraints or legislation are in place (see examples of Figure~\ref{fig:selected_models_tuner_val}).
For Fairband with static $\alpha$, a target trade-off has already been chosen for the method’s search phase, and we once again employ this trade-off for model selection (selection-$\alpha$).
For the FB-auto variant of Fairband, aiming for an automated balance between both metrics, we employ the same strategy for setting $\alpha$ as that used during search.
By doing so, the weight of each metric is pondered by an approximation of their true range instead of blindly applying a pre-determined weight.
For instance, if the distribution of fairness is in range $f \in [0, 0.9]$ but that of accuracy is in range $a \in [0, 0.3]$, then a balance would arguably be achieved by weighing accuracy higher, as each unit increase in accuracy represents a more significant relative change (this mechanism is achieved by Equation~\ref{eq:dynamic_alpha}).
However, at this stage we can use information from all brackets, as we no longer want to promote exploration of the search space but instead aim for a consistent and stable model selection.
Thus, for FB-auto, the selection-$\alpha$ %\footnote{The value used to weigh the two optimization metrics for model selection.}
is chosen from the average fairness and predictive accuracy of all sampled configurations.

% %%%%%%%%%%%%%%%%%%%%%%%%%%%%%%%%%%%%%%%%%%%%%%%%%%%%%%%%%%%%%%%%%%%%%%%%%%%%%%%%%%%%%%%%%%%%%%%%%%%%%%%%%%%%%%%%
% % 
% %                                            EXPERIMENTAL SETUP
% % 
% % %%%%%%%%%%%%%%%%%%%%%%%%%%%%%%%%%%%%%%%%%%%%%%%%%%%%%%%%%%%%%%%%%%%%%%%%%%%%%%%%%%%%%%%%%%%%%%%%%%%%%%%%%%%%%

\section{Experimental Setup}

In order to validate our proposal, we evaluate Fairband on a search space spanning multiple ML algorithms, model hyperparameters, and sampling choices on four different datasets.

We compare our method to several baselines in the hyperparameter optimization community, including Random Search (RS) and Hyperband (HB).
We study two versions of Fairband: FB-auto (employing the dynamic $\alpha$ strategy) and FB-bal (employing $\alpha = 0.5$).
In addition, we consider an RS variant in which we introduce fairness-awareness into the final model selection criteria, giving equal importance to both metrics, \ie, $\alpha=0.5$ (RS-bal).
%In addition, to further investigate the impact of a decision pondered on both fairness and predictive accuracy, we evaluate an RS variant that uses $\alpha=0.5$ for model selection (\textit{RS-bal}).
% This is implemented by selecting the model that better balances both objectives simultaneously.

\subsection{Datasets}

We validate our methodology on three datasets from the fairness literature, and one large-scale case study
%real-world dataset 
on online bank account opening fraud.
Both accuracy and fairness metrics are highly dependent on the task for which a given model is trained, and the real-world setting in which it will be deployed~\citep{bigdatasocialsciences}.
Thus, we detail a task for each of the datasets we employ, subsequently deriving the metrics we use for each.
The \textbf{Donors Choose} dataset~\citep{donorschoose2014} consists in data pertaining to thousands of projects proposed for/by K-12 schools.
The objective is to identify projects at risk of getting underfunded to provide tailored interventions.
As an assistive funding setting, we set a limit of 1000 positive predictions (PP), and select balanced true positive rates across schools from different poverty levels as the fairness metric, also known as equal opportunity~\citep{Hardt2016}.
We use precision as a metric of predictive accuracy.

The \textbf{Adult} dataset~\citep{kohavi1996scaling} consists on data from the 1994 US census, including age, gender, race, occupation, and income, among others.
In order to properly employ this dataset, 
we devise a scenario of a social security program targeting low-income individuals.
In this setting, a positive prediction indicates an income of less than \$50K per year, thus making that person eligible for the assistive program.
As an assistive setting, we select balanced true positive rates across genders as the fairness metric (equal opportunity).
We target a global true positive rate (recall) of 50\%, catching half of the total universe of individuals at need of assistance.
Additionally, aiming to help only those that need it, we use precision as a metric of predictive accuracy.

%% SIZE: 6172
The \textbf{COMPAS} dataset~\citep{Angwin2016} is a criminal justice dataset whose objective is to predict whether someone will re-offend based on the person's criminal history, demographics, and jail time.
As a punitive setting, we select balanced false positive rates for individuals of different races as the fairness metric, also known as predictive equality~\citep{Corbett-Davies2017}.
At the same time, we target a global false positive rate of 2\%, in order to maintain a very low number of unjustly jailed individuals.
% As such, a fair classifier will have similar false positive rates between different races.
%That is, if the classifier is fair, a non-Caucasian individual should \textit{not} have a higher likelihood of being falsely jailed than a Caucasian individual.
Regarding predictive accuracy, we use the precision metric.%, as a well-performing classifier should not ...

The \textbf{AOF} dataset is a large-scale (500K instances) real-world dataset on online bank account opening fraud.
The objective is to predict whether an individual's request for opening a bank account is fraudulent.
As a punitive setting, we select balanced false positive rates across age groups as the fairness metric (predictive equality).
In this setting, a false positive is a genuine customer that sees her/his request unjustly denied, costing the company potential earnings, and disturbing the customer's life.
To maintain a low costumer attrition, we target a global false positive rate of 5\%.
In addition, we use recall (true positive rate) as a metric for predictive accuracy, as we want to promote models that correctly catch a high volume of fraud.
Table~\ref{tab:datasets} summarizes the task details for all datasets.

\begin{table*}[tb]
    \setlength{\tabcolsep}{2pt}
    \centering
    \begin{tabular}{lccccc}
    \toprule
    \textbf{Dataset} & \textbf{Setting} & \textbf{Acc. Metric} & \textbf{Fairness Metric} & \textbf{Target Threshold} & \textbf{Sensitive Attribute} \\
    \midrule
    Donors Choose & assistive & precision & equal opportunity & 1000 PP & poverty level \\
    Adult   & assistive & precision & equal opportunity & 50\% TPR & gender \\
    COMPAS  & punitive  & precision & predictive equality & 2\% FPR & race \\
    AOF     & punitive & recall & predictive equality & 5\% FPR & age \\
    \bottomrule
    \end{tabular}
    \caption{Details on used datasets and their metrics.}
    \label{tab:datasets}
\end{table*}

%% SEARCH SPACE %%
\subsection{Search Space}

First and foremost, we define a hyperparameter as any parameter used to tune an algorithm's learning process~\citep{Hutter2019}.
Both accuracy and fairness metrics are seen as (possibly noisy) black-box functions of these hyperparameters.
This broad definition of hyperparameters includes the model type (which can be abstracted as a categorical hyperparameter), data sampling criteria, as well as regular hyperparameters pertaining only to the model.

In order to validate our methods, we define a comprehensive hyperparameter search space, allowing us to effectively navigate the fairness-accuracy trade-off.
We select five ML model types: Random Forest (RF)~\citep{Breiman2001}, Decision Tree (DT)~\citep{breiman1984classification}, Logistic Regression (LR)~\citep{Walker1967}, LightGBM (LGBM)~\citep{LGBM}, and feed-forward neural networks (NN).
Each model type has the same likelihood of being selected when randomly sampling configurations to test.
When sampling a hyperparameter configuration, after a model type is randomly picked, we take a random sample of that model's hyperparameters.
% For each type, we randomly sample the same number of hyperparameter configurations. 
% Details on the search space for each model type are in the Supplementary Materials.
%
Besides the model type and model hyperparameters, we also experiment with three different undersampling strategies: targeting 20\%, 10\%, and 5\% positive samples.
This type of hyperparameter is only used in the AOF dataset, as its high class imbalance poses a challenge (it features 99 negatively-labeled samples per each positively-labeled sample).
Thus, we define a \textit{model} as the result of training a given \textit{hyperparameter configuration} on a given train dataset.
% Note that a hyperparameter optimization algorithm searches for the optimal \textit{hyperparameter configuration}, not the optimal model.

\subsection{Hyperband Parameters}
\label{sec:hyperband_params}

We configure Fairband to execute SH brackets as in the original Hyperband algorithm.
Following the authors \citep{Li2016b}, we set $\eta = 3$.
%Our method's execution follows the SH brackets defined by the Hyperband algorithm, parameterized with $\eta = 3$, as advised by the authors~\citep{Li2016b}.
% We define 1 budget unit as 1\% of the train dataset, and $R=100$, meaning 100\% of the train dataset is the maximum budget any model is granted.
We define 1 budget unit as 1\% of the train dataset, and we set the maximum budget allocated to any configuration as 100 budget units, $R=100$ (100\% of the training dataset).
These settings result in:

\begin{equation}
     s_{max} = \lfloor\log_{\eta}{(R)}\rfloor = 4, \quad B = (s_{max}+1)\cdot{R} = 500
\end{equation}

The outer loop will run $s_{max} + 1 = 5$ times, for $s \in \{s_{max}, ..., 0\}$.
Each run (or bracket) will consume at most $B=500$ budget units.
Accordingly, each bracket will use at most $s_{max} + 1 = 5$ training slices of increasing size, corresponding to the following dataset percentages: 1.23\%, 3.70\%, 11.1\%, 33.3\%, 100\%.
These training slices are sampled such that smaller slices are contained in larger slices, and such that the class-ratio is maintained (by stratified sampling).
%
% In case of streaming algorithms (\eg, neural networks), this enables the training of all 5 slices in one pass through the data.
% %
% For instance, a streaming model in bracket $s=4$ would consume $100$ budget units to reach the last iteration, instead of $1.23+3.7+11.1+33.3+100=149.33$ budget units\footnote{
% The maximum cumulative budget consumed by a model that reaches the last iteration in a bracket is given by $\frac{\eta}{\eta - 1}$.}.
%
The number of configurations and budget per configuration within each bracket are displayed in Table~\ref{tab:hyperband_brackets}.
Throughout a complete Hyperband run, 143 unique hyperparameter configurations will be randomly sampled (total $n_i$ for $i=0$), and 206 models will be trained and evaluated (total $n_i$ for $i \in \{0 ..., s_{max}\}$).

% \begin{table}[t]    % Keep this table at the top of a page, it's important for future reference!
%     \centering
%     \resizebox{\columnwidth}{!}{
%     \input{Fairband/tables/hyperband_brackets}
%     }
%     \caption{Number of sampled configurations, $n_i$, and budget per configuration, $r_i$, for each Hyperband bracket (when $\eta = 3$ and $R=100$). One budget unit equals 1\% of the train dataset.}
%     \label{tab:hyperband_brackets}
% \end{table}

% %%%%%%%%%%%%%%%%%%%%%%%%%%%%%%%%%%%%%%%%%%%%%%%%%%%%%%%%%%%%%%%%%%%%%%%%%%%%%%%%%%%%%%%%%%%%%%%%%%%%%%%%%%%%%%%%
% % 
% %                                               RESULTS
% % 
% % %%%%%%%%%%%%%%%%%%%%%%%%%%%%%%%%%%%%%%%%%%%%%%%%%%%%%%%%%%%%%%%%%%%%%%%%%%%%%%%%%%%%%%%%%%%%%%%%%%%%%%%%%%%%%
\section{Results \& Discussion}

In this section, we present and analyze the results from our fairness-aware HO experiments.
To validate the methodology, we guide hyperparameter search by evaluating all sampled configurations on the same validation dataset, while evaluating the best-performing configuration (according to the objective function $o$) on a held-out test dataset in the end.
Likewise, the model thresholds are set on the validation dataset, and then used on both the validation and test datasets.
All studied HO methods are given the same training budget: 2400 budget units.
% For RS, this corresponds to training 24 distinct configurations on 100\% of the training dataset.
% For HB and FB, it corresponds to a full run with $\eta=3$ and $R=100$.

% we hold-out a test dataset for each considered dataset.
% % The result of executing a  hyperparameter optimization method is a hyperparameter configuration.
% % We run our proposed methods as well as the literature baselines on 
% Our hyperparameter tuners are guided by results on a validation dataset, model thresholds are set on the same validation dataset, and final performance is assessed on a held-out test dataset.
% The model corresponding to the best hyperparameter configuration on validation data is then evaluated on the held-out test dataset.

\begin{table}[tb]
    \centering
    \input{tables/full_results_table_15runs}
    \caption{
    Validation and test results for all algorithms on all datasets.
    Statistical significance is tested against the baseline models (RS and HB) with a Kolmogorov-Smirnov test~\citep{doi:10.1080/01621459.1967.10482916}.
    Comparison with RS is indicated by $\lozenge$ when $p < 0.05$ ($\blacklozenge$ when $p < 0.01$), and comparison with HB is indicated by $\vartriangle$ when $p < 0.05$ ($\blacktriangle$ when $p < 0.01$).
    }
    \label{tab:full_results_table}
\end{table}

% \subsection{General Results} 
Table~\ref{tab:full_results_table} shows the validation and test results of running different HO methods on four chosen datasets\footnote{Data, plots, and ML artifacts available at \url{https://github.com/feedzai/fair-automl}}.
Results are averaged over 15 runs, and statistical significant differences against the baseline methods are shown with $\vartriangle$ (HB) and $\lozenge$ (RS).
This table comprises the training and evaluation of 40K unique models, one of the largest studies of the fairness-accuracy trade-off to date.
Our method (FB-auto) consistently achieves higher fairness than the baselines on all datasets, at a small cost in predictive accuracy (statistically significant on all datasets).
The same trend is observed with the remaining fairness-aware methods (FB-bal and RS-bal), when compared with the fairness-blind methods (RS and HB).
%

%%%%%%%%%%%%%%%%%%%%%%%%%%%
%% Analysis Dataset-wise %%
%%%%%%%%%%%%%%%%%%%%%%%%%%%

The differences in predictive accuracy and fairness between the proposed fairness-aware methods and the fairness-blind baselines have strong statistical significance on the Donors Choose, Adult, and AOF datasets.
The same trend is visible on the COMPAS dataset, although on this dataset fairness-blind RS does not achieve better performance than FB, while achieving substantially lower fairness.
%
% On the AOF dataset, all fairness-aware methods achieve significantly higher fairness than their fairness-blind counterparts, while differences in performance are not significant for the pairs RS-bal--RS, RS-bal--HB, and FB-auto--RS.
%
%%% ON COMPAS
%
We note that the COMPAS dataset is the smallest by a significant margin (approximately one order of magnitude smaller than Adult and Donors Choose, and two orders smaller than AOF).
We find the large variability in fairness between validation and test results on COMPAS to be best explained by its size, together with the strict target threshold of 2\% FPR (which is set on the validation data, and used on both validation and test data).
%
% Furthermore, we observe that while the obtained accuracy in the holdout test set is on a par with the values on validation, the same does not hold for fairness, whose values drop abruptly. We find this behavior to be best explained by the threshold and the dataset size. As COMPAS is comparatively small, setting models' threshold on a validation set (approximately 20\% of the dataset) is very sensitive to the actual data.
%

%%%%%%%%%%%%%%
%% OVERALL  %%
%%%%%%%%%%%%%%

\begin{table}[tbh]
    \centering
    \begin{tabular}{lcccc}
\toprule
\multirow{2}{*}{\textbf{Dataset}} & \multicolumn{2}{c}{\textbf{Abs. Difference (pp)}} & \multicolumn{2}{c}{\textbf{Rel. Difference (\%)}}   \\
     & \textbf{Predictive Acc.} & \textbf{Fairness} & \textbf{Predictive Acc.} & \textbf{Fairness} \\
\midrule
    Donors Choose   & -3.4 & +52.6 & -6.4 & +152 \\
    Adult           & -9.3 & +40.6 & -9.4 & +76.2 \\
    COMPAS          & -3.4 & +17.5 & -4.1 & +72.6 \\
    AOF             & -6.4 & +31.0 & -9.3 & +70.6 \\
\midrule
    \textit{Average}   & -5.6  & +35.4 & -7.3 & +92.9 \\
\bottomrule
\\
    \end{tabular}
    \caption{Comparison of using Fairband (FB-auto) versus Hyperband (HB), on test results.}
    \label{tab:fb_hb_comparison}
\end{table}

Overall, FB-auto arguably achieves the best fairness-accuracy trade-off on three (Donors Choose, Adult, AOF) out of the four datasets.
On the COMPAS dataset, FB-bal dominates the remaining fairness-aware methods (although differences to FB-auto are not statistically significant).
% Indeed, its solutions dominate the remaining fairness-aware methods (FB-bal and RS-bal) on Adult and COMPAS, achieving simultaneous fairness and accuracy improvements on the test data.
%
Unsurprisingly, HB achieves the highest predictive accuracy on all datasets.
However, by using FB-auto we achieve 92.9\% improvement in fairness at a cost of only 7.3\% drop in predictive accuracy, averaged over all datasets.
Table~\ref{tab:fb_hb_comparison} summarizes the comparison between HB and FB-auto on all datasets.

\begin{figure}[H]
\begin{multicols}{2}

\begin{figure}[H]
    \centering
       \includegraphics[width=0.7\columnwidth]{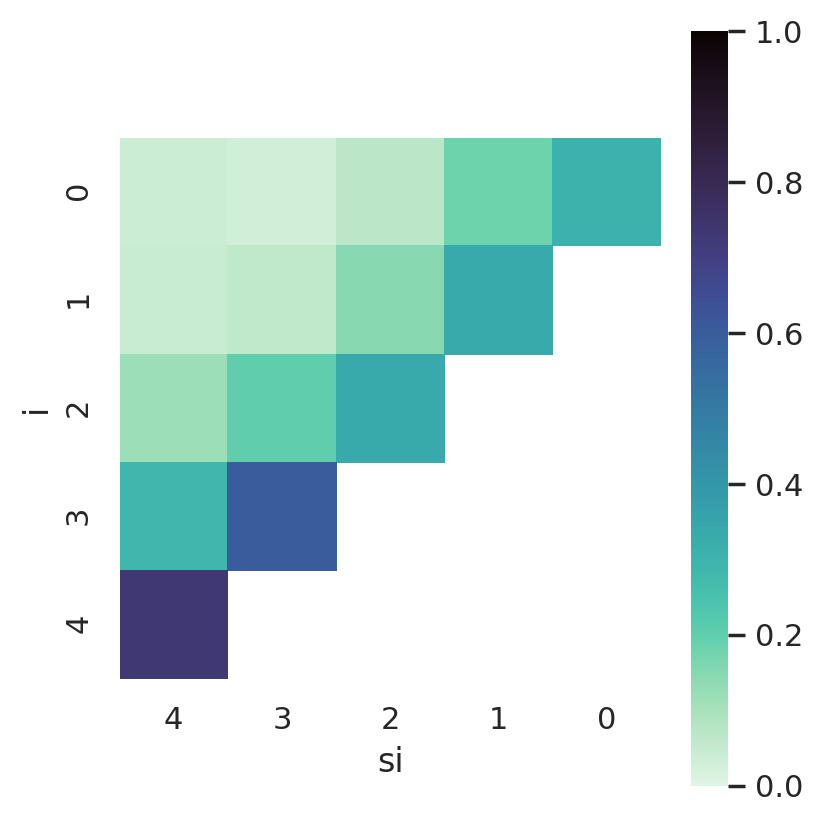}
\end{figure}

\begin{figure}[H]
    \centering
        \includegraphics[width=0.7\columnwidth]{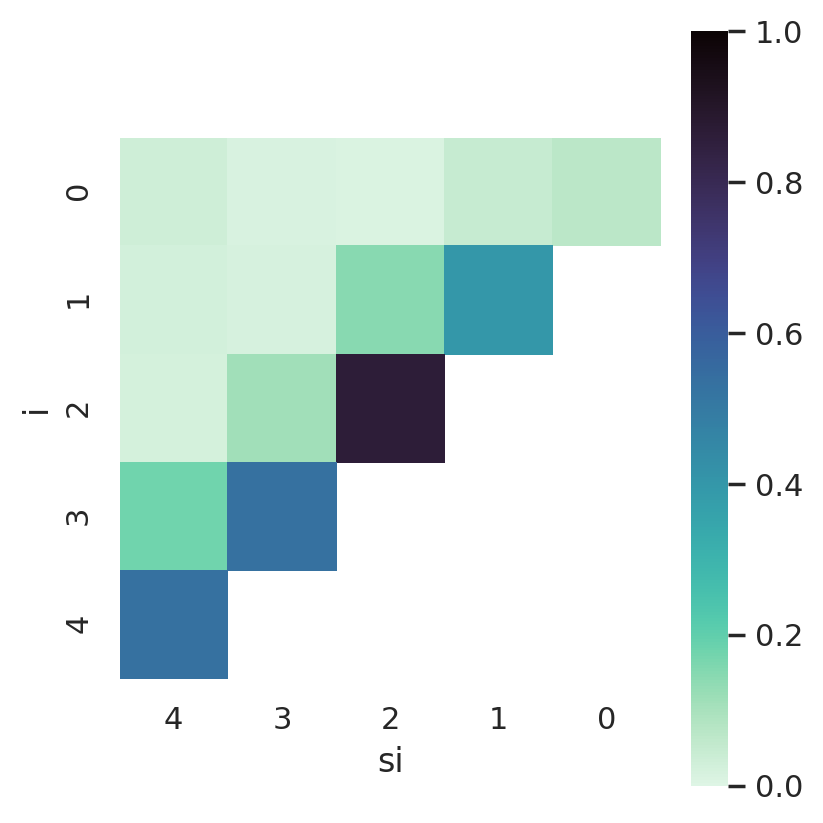}
\end{figure}
\end{multicols}

\caption{Average density of Pareto optimal models per FB-auto iteration (Adult dataset on left plot, AOF on right plot). Refer to Table~\ref{tab:hyperband_brackets} for information on the configurations at each iteration.}
\label{fig:heatmaps}

\end{figure}

\begin{figure}[H]
\begin{multicols}{2}

\begin{figure}[H]
    \centering
       \includegraphics[width=1.0\columnwidth]{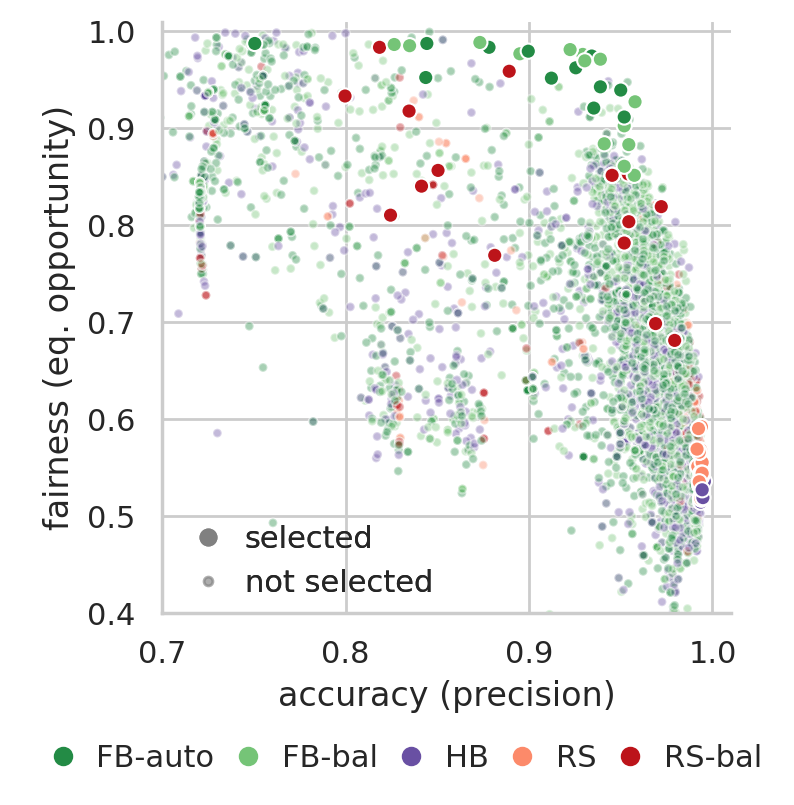}
    % \caption{Caption}
    % \label{fig:my_label}
\end{figure}

\begin{figure}[H]
    \centering
        \includegraphics[width=1.0\columnwidth]{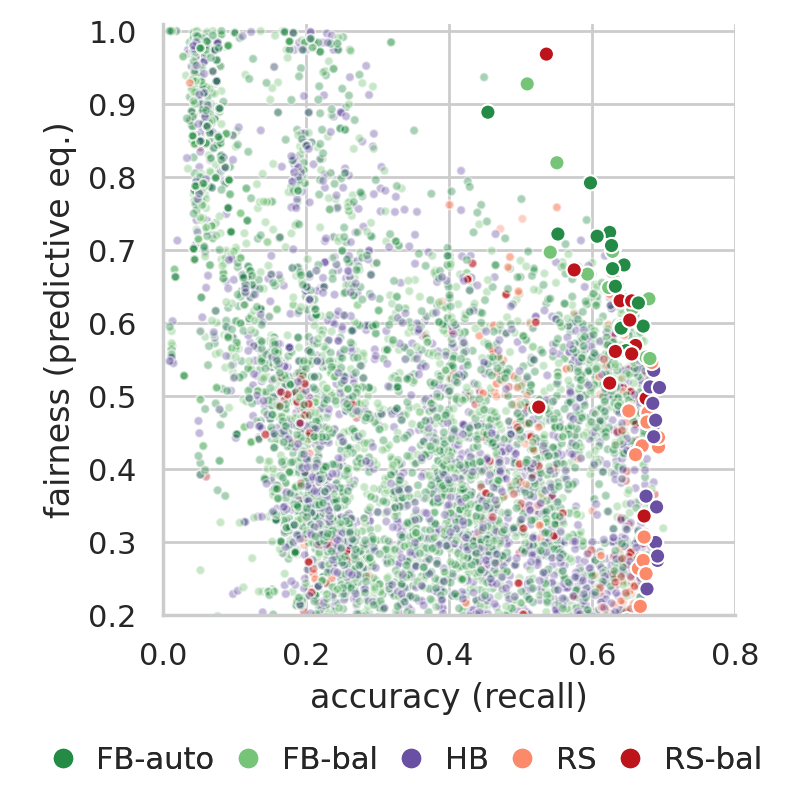}
    % \caption{Caption}
    % \label{fig:my_label}
\end{figure}
\end{multicols}

\caption{Fairness and predictive accuracy of selected models by hyperparameter optimization algorithm (Adult dataset on left plot, AOF on right plot).}
    \label{fig:selected_models_tuner_val}

\end{figure}

% Search Strategy
\subsection{Search Strategy} 
\label{sssec:search_strategy}

We evaluate the search strategy by analyzing the evolution of fairness and performance simultaneously, and whether we can effectively extend the practical Pareto Frontier as the search progresses.
That is, whether optimal trade-offs are more likely to be found as we discard the worst performing models and further increase the allocated budget for top-performing models within each bracket (see Table~\ref{tab:hyperband_brackets}).
%
%In order to evaluate our method's search strategy, we must study the evolution of fairness and performance simultaneously, and whether we can effectively extend the practical Pareto frontier as the search progresses.
%
% Figure~\ref{fig:heatmaps} shows a heat map of the density of Pareto optimal models\footnote{Fraction of Pareto optimal models within each bracket, with optimality assessed within each run.} in each FB iteration, for FB-auto, on the datasets Adult and AOF
% \footnote{Similar results are observed on the other datasets.}.
Figure~\ref{fig:heatmaps} shows a heat map of the average density of Pareto optimal models\footnote{Fraction of Pareto optimal models within each bracket, with optimality assessed within each run.} in each FB iteration, for 15 runs of the FB-auto algorithm, on the Adult and AOF datasets. %(similar results are observed on all datasets).
As the iterations progress, under-performing configurations are pruned, and the density of Pareto optimal models steadily increases, confirming the effectiveness of the search strategy.
% The continuous increase in the density of Pareto optimal models as the iterations progress and under-performing configurations are pruned, confirms the effectiveness of the search strategy.
%
%An effective search strategy will continuously increase the density of Pareto optimal models as the iterations progress, and as under-performing configurations are pruned.
%
%This trend is clearly visible on all datasets (only Adult and AOF shown).

% An effective search strategy is expected to continuously increase the density of Pareto optimal models as the iteration $i$ progresses. This can be observed on both plots by the low density of optimal models on $i=0$, and increasingly high density in later iterations. 

%%%% -----------------------------------
%%%         MODEL SELECTION 
%%%% -----------------------------------
\subsection{Model Selection}

Figure~\ref{fig:selected_models_tuner_val} shows the final selected models for 15 runs of each method, respectively for the Adult (left) the AOF (right) datasets.
The remaining models considered during the search are also shown, with lower opacity and smaller size.
%All sampled models during the methods' execution are also shown, with lower opacity and smaller size.
As can be seen by the plots, Fairband consistently identifies good fairness-accuracy trade-offs from the universe of available configurations.
Moreover, Fairband often achieves higher fairness than RS-bal for the same predictive accuracy.
Indeed, the models selected by Fairband are consistently close to or form the Pareto frontier.
Most importantly, as evident by the spread of selected models, we can successfully navigate the fairness-accuracy Pareto frontier solely by means of HO.

\subsection{Efficiency over Budget}
% \subsection{Method's Efficiency}
% \subsection{Results as Budget Increases}

\begin{figure}[H]
\centering
\includegraphics[width=0.9\columnwidth]{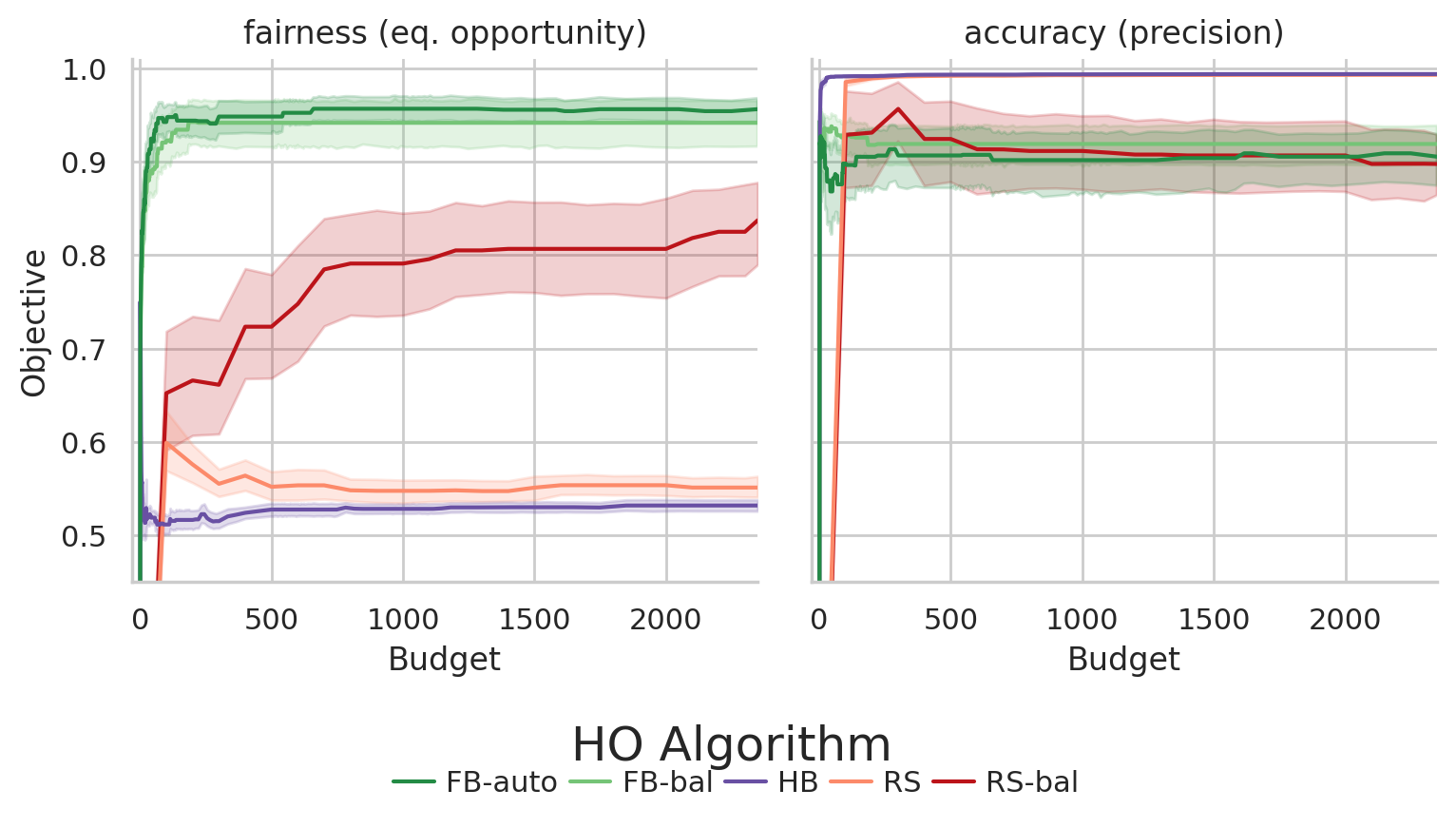}
\caption{Fairness and predictive accuracy as a function of budget in the Adult validation set.}
\label{fig:adult_budget_val_set}
\end{figure}

\begin{figure}[H]
\centering
\centering
\includegraphics[width=0.9\columnwidth]{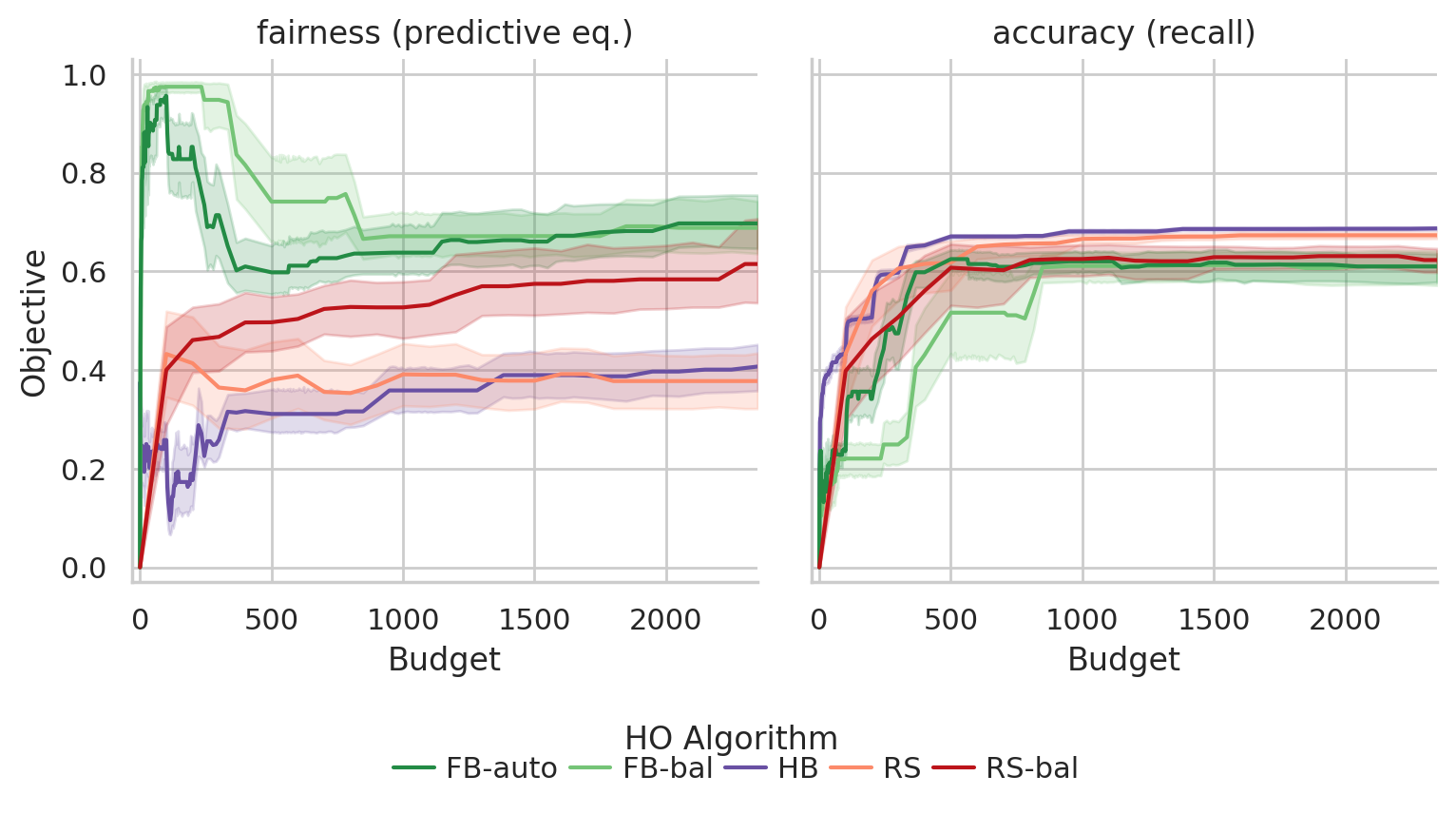}
\caption{Fairness and predictive accuracy as a function of budget in the AOF validation set.}
\label{fig:aof_budget_val_set}
\end{figure}

A different perspective of the HO methods' execution is their progression as the budget increases.
We expect that increasingly better trade-offs are found as the budget increases, guided by our method's $\alpha$.
% We expect the proposed method to find better trade-offs as budget increases as guided by $\alpha$.
Figures~\ref{fig:adult_budget_val_set} and~\ref{fig:aof_budget_val_set} show this progression for the Adult and AOF datasets, respectively (error bands show 95\% confidence intervals)\footnote{Plots for all datasets are available at \url{https://github.com/feedzai/fair-automl}}.

Across all datasets, Fairband (both FB-auto and FB-bal) is able to provide strong anytime fairness, quickly converging to fairer regions of hyperparameter space, while RS-bal finds fairer configurations only at later stages.
Indeed, FB-auto achieves better fairness (difference is statistically significant) at no cost to predictive accuracy (difference is not statistically significant) when compared to RS-bal.
%
% As can be seen in the plots, both FB-bal and FB-auto exhibit similar validation performance. However, the former yields smoother results, as expected, given that $\alpha$ is constant.
%

On the AOF dataset, both versions of Fairband show an acute drop in fairness accompanied by symmetric increase in predictive accuracy by the 500 budget mark.
This initial budget allocation corresponds to the left-most bracket ($s=4$ in Table~\ref{tab:hyperband_brackets}), which is highly exploratory (samples 81 different hyperparameter configurations) and thus trains each configuration on a small initial budget (1.2\% of the training dataset). As configurations are pruned and the budget per configuration increases, the discovered trade-offs are progressively more accurate but less fair.
This steep increase in the training budget per configuration on the first 500 budget units leads to the visibly high variability in objective metrics.
%
%
% In the case of the AOF dataset, both FB-bal and FB-auto present an acute drop in fairness, accompanied by a steep increase in predictive accuracy up until the 500 budget mark, which corresponds to the budget allocated to the left-most bracket in Table~\ref{tab:hyperband_brackets}. This variability is best explained by the initially low budgeted models (each up to 1.2\% of the training set), together with the need to balance both performance metrics (using $\alpha$) as we decide which models to discard in the next iteration.
% %However, the former yields smoother results inherent to the choice of a static $\alpha$ value.

Regarding fairness-blind methods, RS and HB show similar behavior, with predictive accuracy increasing asymptotically, typically at the cost of fairness.
However, HB consistently achieves higher predictive accuracy than RS at all stages.

When compared to literature HO baselines, both versions of Fairband achieve significantly improved fairness at a comparatively small cost in predictive accuracy.
This trend is repeated by RS-bal, although achieving lower fairness than Fairband.
It is important to consider that these fairness-blind baseline methods are the current standard in HO.
By unfolding the fairness dimension, we show that strong predictive accuracy carries an equally strong real-world cost in unfairness, and this is hidden by traditional HO methods.
% By unfolding the fairness dimension, we show that strong predictive accuracy might yield a wide range of fairness values, and this is hidden by traditional HO methods.

%%%%%%%%%%%%%%%%%%%%%%%%%%%%%%%%%%%%%%%
% FAIRLEARN EXPERIMENT
%%%%%%%%%%%%%%%%%%%%%%%%%%%%%%%%%%%%%%%


\subsection{Optimizing Bias Reduction Hyperparameters}
As a hyperparameter optimization method, one fruitful approach to bias-mitigation is adding bias reduction methods to Fairband's search space.
As such, we introduce the Exponentiated Gradient (EG) reduction for fair classification algorithm\footnote{Implemented on the open-source \textit{Fairlearn} package.}~\citep{Agarwal2018}
into our search space on the Adult dataset.
EG is a state-of-the-art bias reduction algorithm that optimizes predictive accuracy subject to fairness constraints, and is compatible with any cost-sensitive binary classifier.
In our setting, we target equal opportunity, and apply EG over a Decision Tree classifier.

\begin{figure}[H]
    \centering
    \includegraphics[width=0.60\columnwidth]{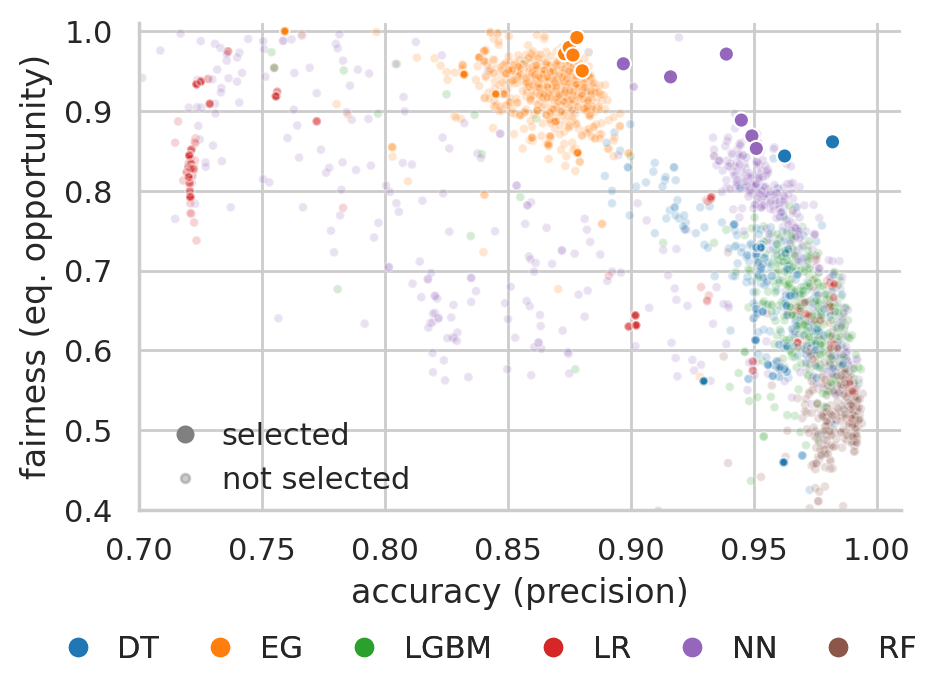}
    \caption{Fairness and predictive accuracy of selected models for FB-auto, discriminated by model type, on the Adult dataset.}
    \label{fig:fairlearn}
\end{figure}

Figure~\ref{fig:fairlearn} shows a plot of the models selected by FB-auto over 15 runs on the Adult dataset.
The introduction of EG creates a new cluster of models in our search space (shown in orange), consisting of possible fairness-accuracy trade-offs in a previously unoccupied region (compare with Figure~\ref{fig:selected_models_tuner_val}, left plot). However, even though these models were trained specifically targeting our fairness metric (equality of opportunity) while the remaining models were trained in a fairness-blind manner, Fairband chooses other model types more often than not.
Indeed, the selected NNs and DTs arguably represent the best fairness-accuracy trade-offs.

Essentially, these results show that blindly applying bias reduction techniques may lead to sub-optimal fairness-accuracy trade-offs. %that can be dominated even by fairness-blind classifiers with the proper hyperparameters.
% Nevertheless, bias reduction methods should also be included in the portfolio of algorithms to be evaluated in fairness-aware model selection processes, as a large selection of models is critical to properly explore the fairness-accuracy search space.
%
Overall, these results support the fact that Fairband should be employed in all ML pipelines that aim for fair decision-making, together with bias-blind methods and bias reduction methods alike, to properly explore the fairness-accuracy search space.
%
% Overall, these results reinforce the claim that Fairband should be employed in all real-world ML pipelines, together with bias reduction methods and bias-blind methods alike.

% %%%%%%%%%%%%%%%%%%%%%%%%%%%%%%%%%%%%%%%%%%%%%%%%%%%%%%%%%%%%%%%%%%%%%%%%%%%%%%%%%%%%%%%%%%%%%%%%%%%%%%%%%%%%%%%%
% % 
% %                                               CONCLUSION
% % 
% % %%%%%%%%%%%%%%%%%%%%%%%%%%%%%%%%%%%%%%%%%%%%%%%%%%%%%%%%%%%%%%%%%%%%%%%%%%%%%%%%%%%%%%%%%%%%%%%%%%%%%%%%%%%%%
 \section{Conclusion}
 \label{sec:conclusions}

There have been widespread reports of real-world AI systems shown to be biased, causing serious disparate impact across different sub-groups, unfairly affecting people based on race, gender or age. The AI research community has embraced this issue and has been doing extensive research work. However, the current landscape of algorithmic fairness lacks (1) practical methodologies and (2) tools for real-world practitioners.

This work aims to bridge that gap by providing a simple and flexible hyperparameter optimization technique to foster the incorporation of fairness objectives in real-world ML pipelines. Fairband is a bandit-based fairness-aware hyperparameter optimization method that extends the Successive Halving and Hyperband algorithms by guiding the hyperparameter search towards fairer configurations\footnote{Fairband is completely agnostic to the hyperparameter search space and therefore it is subject to the fairness-accuracy trade-offs the input hyperparameter configurations are able to attain.}.

Fairband enables targeting a specific fairness-accuracy trade-off (by means of an $\alpha$ parameter), which is often dictated by business restrictions or regulatory law.
Aiming for a complete out-of-the-box experience, we alternatively propose an algorithm for setting $\alpha$ automatically, eliminating the need to tune this parameter.

By introducing fairness notions into hyperparameter optimization, our method can be seamlessly integrated into real-world ML pipelines, at no extra training cost.
Moreover, our method is easy to implement, resource-efficient, and both model- and metric-agnostic, providing no obstacles to its widespread adoption. 

We evaluate our method on four real-world decision-making datasets, and show that it is able to provide significant fairness improvements at a small cost in predictive accuracy, when compared to traditional HO techniques.
%When compared to Hyperband, our method achieves an average of 92.9\% improvement in fairness, at a cost of only 7.3\% decrease in predictive accuracy.
We show that it is both possible and effective to navigate the fairness-accuracy trade-off through hyperparameter optimization.

Crucially, we observe that there is a wide spread of attainable fairness values at any level of predictive accuracy.
At the same time, we once again document the known inverse relation between fairness and predictive accuracy.
Hence, by blindly optimizing a single predictive accuracy metric (as is standard practise in real-world ML systems) we are inherently targeting unfairer regions of the hyperparameter space.

\section*{Acknowledgements}

The project CAMELOT (reference POCI-01-0247-FEDER-045915) leading to this work is co-financed by the ERDF - European Regional Development Fund through the Operational Program for Competitiveness and Internationalisation - COMPETE 2020, the North Portugal Regional Operational Program - NORTE 2020 and by the Portuguese Foundation for Science and Technology - FCT under the CMU Portugal international partnership.

\bibliographystyle{apalike}
\bibliography{refs}

\end{document}

%% file: tables/hyperband_brackets.tex
\begin{tabular}{c|cc|cc|cc|cc|cc}
% \fontsize{9}{11}\selectfont
& \multicolumn{2}{c|}{$s = 4$} & \multicolumn{2}{c|}{$s = 3$} & \multicolumn{2}{c|}{$s = 2$} & \multicolumn{2}{c|}{$s = 1$} & \multicolumn{2}{c}{$s = 0$} \\
$i$	& $n_i$ & $r_i$		& $n_i$ & $r_i$		& $n_i$ & $r_i$ & $n_i$ & $r_i$ & $n_i$ & $r_i$ \\ \hline
0	& 81    & 1.2		& 34    & 3.7		& 15 & 11     & 8 & 33	& 5 & 100  \\
1	& 27    & 3.7		& 11    & 11		& 5  & 33     & 2 & 100	&   &   \\
2	& 9     & 11		& 3     & 33		& 1  & 100      &   & 		&   &   \\
3	& 3     & 33		& 1     & 100		&	 &          &   & 		&   &   \\
4	& 1     & 100		&       & 			&	 &          &   & 		&   &   \\
\end{tabular}

%% file: tables/full_results_table_15runs.tex
\begin{tabular}{lcccc}

\toprule
     \multirow{2}{*}{\textbf{Algo.}} & \multicolumn{2}{c}{\textbf{Validation}}       & \multicolumn{2}{c}{\textbf{Test}}            \\
     & \textbf{Predictive Acc.} & \textbf{Fairness} & \textbf{Predictive Acc.} & \textbf{Fairness} \\
\midrule
%^{\blacktriangle\blacklozenge}
  \multicolumn{5}{c}{Donors Choose} \\
\midrule
  FB-auto &   $ 53.8^{\blacktriangle\blacklozenge} $ &   $ 97.9^{\blacktriangle\blacklozenge} $ &   $ 50.0^{\blacktriangle\blacklozenge} $ &   $ \textbf{87.3}^{\blacktriangle\blacklozenge} $ \\
   FB-bal &   $ 52.8^{\blacktriangle\blacklozenge} $ &   $ \textbf{98.6}^{\blacktriangle\blacklozenge} $ &   $ 49.9^{\blacktriangle\blacklozenge} $ &   $ 86.1^{\blacktriangle\blacklozenge} $ \\
   RS-bal &   $ 52.3^{\blacktriangle\blacklozenge} $ &   $ 95.2^{\blacktriangle\blacklozenge} $ &   $ 50.4^{\blacktriangle\blacklozenge} $ &   $ 82.2^{\blacktriangle\blacklozenge} $ \\
       RS &   $ 59.9 $ &   $ 26.8 $ &   $ 53.1 $ &   $ 32.8 $ \\
       HB &   $ \textbf{60.6} $ &   $ 28.8 $ &   $ \textbf{53.4} $ &   $ 34.7 $ \\
\midrule

  \multicolumn{5}{c}{Adult} \\
\midrule
  FB-auto &   $ 90.6^{\blacktriangle\blacklozenge} $ &   $ \textbf{95.6}^{\blacktriangle\blacklozenge} $ &   $ 90.1^{\blacktriangle\blacklozenge} $ &   $ \textbf{93.9}^{\blacktriangle\blacklozenge} $ \\
   FB-bal &   $ 91.9^{\blacktriangle\blacklozenge} $ &   $ 94.2^{\blacktriangle\blacklozenge} $ &   $ 79.7^{\blacktriangle\blacklozenge} $ &   $ 79.0^{\blacktriangle\blacklozenge} $ \\
   RS-bal &   $ 89.8^{\blacktriangle\blacklozenge} $ &   $ 83.7^{\blacktriangle\blacklozenge} $ &   $ 90.5^{\blacktriangle\blacklozenge} $ &   $ 83.4^{\blacktriangle\blacklozenge} $ \\
       RS &   $ 99.3 $ &   $ 55.1 $ &   $ \textbf{99.4} $ &   $ 55.1 $ \\
       HB &   $ \textbf{99.4} $ &   $ 53.2 $ &   $ \textbf{99.4} $ &   $ 53.3 $ \\
\midrule

  \multicolumn{5}{c}{COMPAS} \\
\midrule
  FB-auto &   $ 83.9^{\blacktriangle\lozenge} $ &   $ \textbf{97.0}^{\blacktriangle\blacklozenge} $ &   $ 79.2^{\blacktriangle} $ &   $ 41.6^{\blacktriangle} $ \\
   FB-bal &   $ 84.2^{\blacktriangle\lozenge} $ &   $ 96.0^{\blacktriangle\blacklozenge} $ &   $ 79.4^{\blacktriangle} $ &   $ \textbf{42.7}^{\blacktriangle} $ \\
   RS-bal &   $ 80.8^{\blacktriangle\blacklozenge} $ &   $ 77.2^{\blacktriangle\blacklozenge} $ &   $ 74.8^{\blacktriangle\lozenge} $ &   $ 40.7^{\blacktriangle\blacklozenge} $ \\
       RS &   $ 86.8 $ &   $ 29.0 $ &   $ 79.7 $ &   $ 25.8 $ \\
       HB &   $ \textbf{88.6} $ &   $ 17.7 $ &   $ \textbf{82.6} $ &   $ 24.1 $ \\
\midrule

  \multicolumn{5}{c}{AOF} \\
\midrule
  FB-auto &   $ 61.0^{\blacktriangle\blacklozenge} $ &   $ \textbf{69.7}^{\blacktriangle\blacklozenge} $ &   $ 62.6^{\blacktriangle\blacklozenge} $ &   $ \textbf{74.9}^{\blacktriangle\blacklozenge} $ \\
   FB-bal &   $ 61.0^{\blacktriangle\blacklozenge} $ &   $ 68.8^{\blacktriangle\blacklozenge} $ &   $ 62.1^{\blacktriangle\blacklozenge} $ &   $ 72.9^{\blacktriangle\blacklozenge} $ \\
   RS-bal &   $ 62.3^{\blacktriangle\blacklozenge} $ &   $ 61.5^{\blacktriangle\blacklozenge} $ &   $ 63.8^{\blacktriangle} $ &   $ 66.6^{\blacktriangle\blacklozenge} $ \\
       RS &   $ 67.3 $ &   $ 37.8 $ &   $ 67.4 $ &   $ 40.4 $ \\
       HB &   $ \textbf{68.7} $ &   $ 40.7 $ &   $ \textbf{69.0} $ &   $ 43.9 $ \\
\bottomrule
\\
\end{tabular}